\documentclass{article} 
\usepackage{iclr2021_conference,times}

\usepackage{amsmath,amsfonts,bm}

\def\eqref#1{equation~\ref{#1}}

\def\1{\bm{1}}

\DeclareMathAlphabet{\mathsfit}{\encodingdefault}{\sfdefault}{m}{sl}
\SetMathAlphabet{\mathsfit}{bold}{\encodingdefault}{\sfdefault}{bx}{n}

\usepackage[colorlinks=true,linkcolor=blue,citecolor=black]{hyperref}
\usepackage{url}

\usepackage{caption}
\usepackage{graphicx,subcaption}
\usepackage{amsmath}
\usepackage{amssymb}
\usepackage{multirow}
\newcommand{\STAB}[1]{\begin{tabular}{@{}c@{}}#1\end{tabular}}
\usepackage{booktabs}
\usepackage{xcolor}
\usepackage{paralist}

\title{RobustPointSet: A Dataset for Benchmarking Robustness of Point Cloud Classifiers}

\author{Saeid Asgari Taghanaki \& Jieliang Luo \thanks{Equal contribution; corresponding author: \texttt{saeid.asgari.taghanaki@autodesk.com}} \\
Autodesk AI Lab \\
\AND
Ran Zhang, Ye Wang \& Pradeep Kumar Jayaraman \\
Autodesk Research \\
\AND
Krishna Murthy Jatavallabhula \\
Mila, Universit\'e de Montr\'eal \\
}

\iclrfinalcopy 
\begin{document}

\maketitle

\begin{abstract}
The 3D deep learning community has seen significant strides in pointcloud processing over the last few years. However, the datasets on which deep models have been trained have largely remained the same.
Most datasets comprise clean, clutter-free pointclouds canonicalized for pose. Models trained on these datasets fail in uninterpretible and unintuitive ways when presented with data that contains transformations ``unseen'' at train time.
While data augmentation enables models to be robust to ``previously seen'' input transformations, 1) we show that this does not work for unseen transformations during inference, and 2) data augmentation makes it difficult to analyze a model's inherent robustness to transformations. To this end, we create a publicly available dataset for robustness analysis of point cloud classification models (independent of data augmentation) to input transformations, called \textbf{RobustPointSet}.
Our experiments indicate that despite all the progress in the point cloud classification, there is no single architecture that consistently performs better---several fail drastically---when evaluated on transformed test sets. We also find that robustness to unseen transformations cannot be brought about merely by extensive data augmentation. RobustPointSet can be accessed through \url{https://github.com/AutodeskAILab/RobustPointSet}. 
\end{abstract}

\section{Introduction}

The performance of deep neural networks is often measured by their predictive behavior on a test set. However, evaluating a model on an independently and identically distributed (i.i.d.) test set fails to capture its underlying behavior even if the dataset is large enough~\cite{geirhos2020shortcut}. Despite the growing interest in robustness analysis of 2D deep models, less effort has been made in studying the robustness of the models processing point clouds, i.e., sparse permutation-invariant point sets representing 3D geometric data. 

\begin{figure} []
    \centering
    \begin{subfigure}[b]{0.35\linewidth}        
        \centering
        \includegraphics[width=\linewidth]{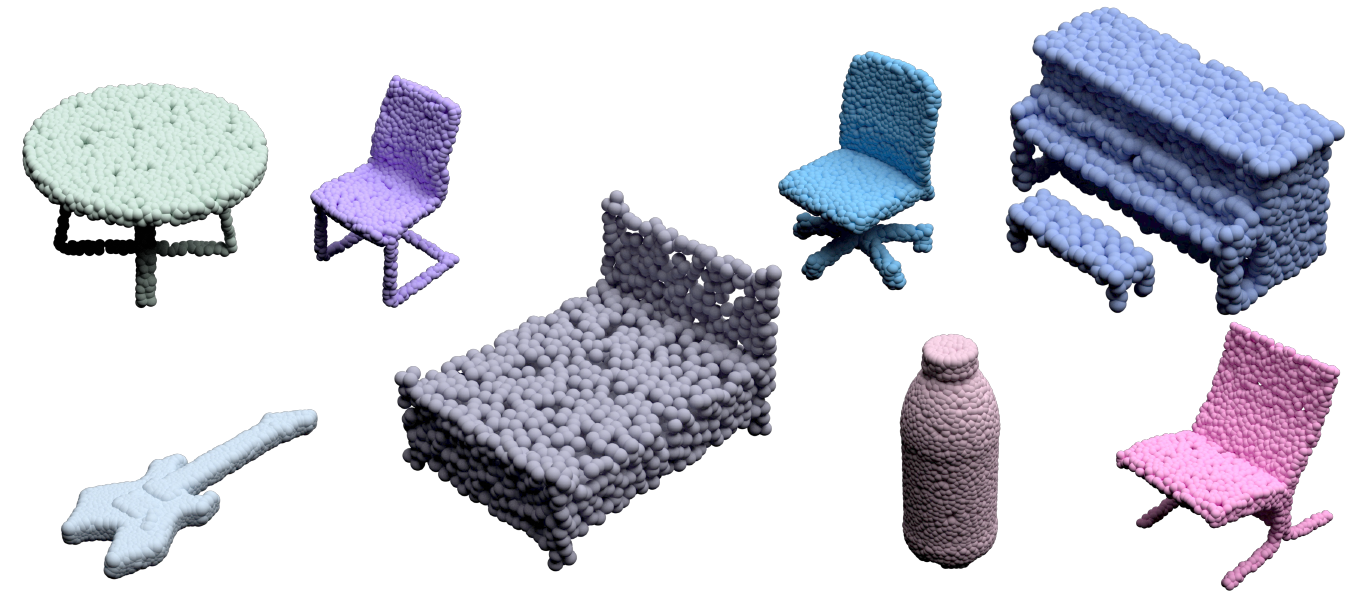}
        \caption{Original}
        \label{fig:A}
    \end{subfigure}\hspace{2cm}
    \begin{subfigure}[b]{0.35\linewidth}        
        \centering
        \includegraphics[width=\linewidth]{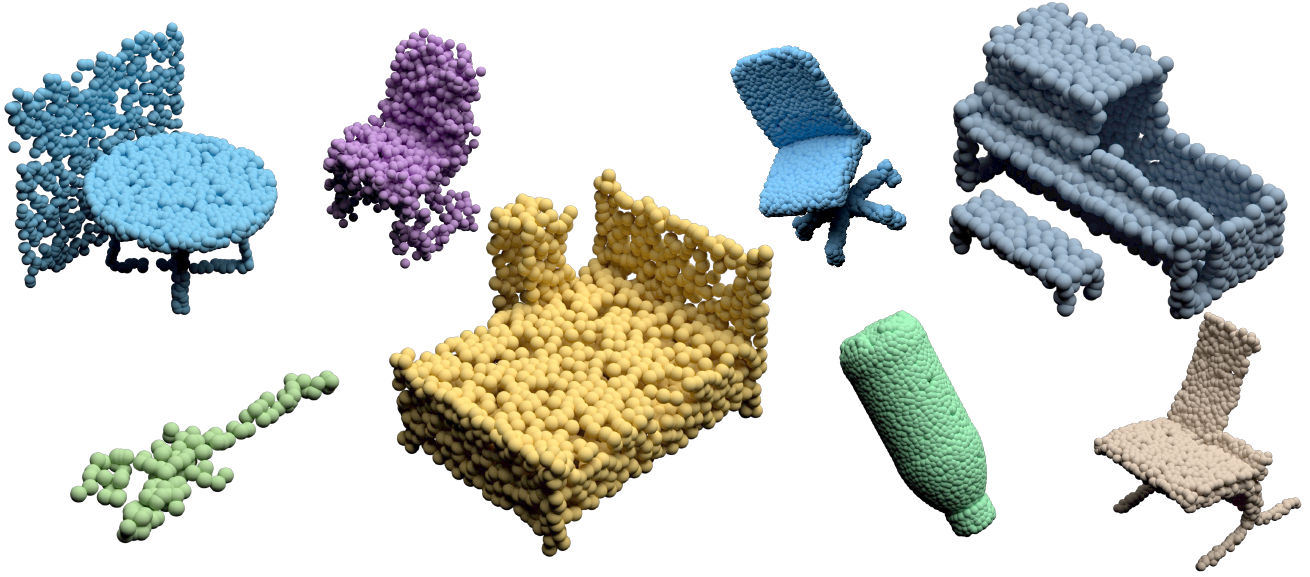}
        \caption{Transformed}
        \label{fig:B}
    \end{subfigure}
    \caption{Pointcloud classification models trained on clean 3D datasets (a) often fail on data with ``unseen" transformations (b). We thoroughly analyze several pointcloud processing architectures and observe that none of them consistently outperform the others. To further research in the robustness of pointcloud classifiers, we created \textbf{\textit{RobustPointSet}}, a publicly available dataset, and specified evaluation protocols to quantify the robustness of models to ``unseen'' transformations during training.}
    \label{fig:teaser}
    \vspace{-0.4cm}
\end{figure}

In practice, neural networks are often deployed to work with real-world pointcloud data that are likely to be transformed in several ways. Point clouds obtained from 3D scans could be corrupted by sensor noise, parts of the object may be missing due to occlusions, the object could be in a different translated or rotated coordinate system, etc. (Figure~\ref{fig:teaser}).
Although data augmentation might help a model to perform well facing a \textit{previously seen} input transformation, it does not improve the performance of a model on handling different \textit{unseen} transformations. Moreover, it hides the real performance of a model processing transformed inputs e.g., rotation augmentation improves a model's performance on unseen rotated objects, however, this does not indicate whether the model itself is rotation-invariant.

In order to take a step towards critical analysis and study the robustness of point cloud classification models \textit{independent} of data augmentation, we created a benchmark dataset called RobustPointSet which is a variant of ModelNet40 dataset~\cite{wu20153d}. In addition to the original training and test sets of ModelNet40, our RobustPointSet consists of six different test sets each with a plausible transformation. 

In this paper, we make the following contributions:

\begin{compactenum}
    \setlength\itemsep{0.5pt}
    \item We create a benchmark dataset for \textit{augmentation-independent} robustness analysis of point cloud classification models.
    \item We empirically demonstrate that despite achieving high classification accuracy on aligned and noise free objects, there is still a large space for improving point cloud classification models to be able to operate well on challenging test samples.
    \item We show that data-augmentation has a very small effect on unseen transformations.
\end{compactenum}

\section{Related Work}
Few robust methods exist for handling transformed point clouds during inference.
Recently, Yan et al.~\cite{yan2020pointasnl} proposed an adaptive sampling scheme to compute non-local features to make the model robust to noise.
While this method takes a critical step towards robustness analysis, it is mainly tailored to handle noise.
Xiao et al~\cite{xiao2020trianglenet} presented a descriptor that is insensitive to pose and sparsity.
However, it is not clear if their method generalizes to other unseen transformations during inference.

There are a limited number of datasets that can be used to train point cloud classification methods. A common dataset is ModelNet40~\cite{wu20153d} which contains 12,308 CAD models of 40 categories. ShapeNet~\cite{chang2015shapenet} is another large-scale dataset of 3D CAD shapes with approximately 51,000 objects in 55 categories. However, in these datasets objects are aligned, complete, and free from any background noise which often is not the case in practice.  ScanObjectNN~\cite{uy2019revisiting} is an alternative relatively small dataset with 2,902 samples collected from scanned indoor scene data that are grouped into 15 classes. ScanObjectNN puts together objects from SceneNN~\cite{hua2016scenenn} and ScanNet~\cite{dai2017scannet} datasets to cover some of the real-world challenges such as background noise and incomplete objects with the goal of analyzing the general performance of point cloud methods in more challenging scenarios. However, as a mix of multiple transformations might happen in both training and test sets of ScanObjectNN, it is impossible to study models' behaviors on particular transformations separately.

\section{RobustPointSet}
\begin{figure*}[ht!]
     \centering
     \includegraphics[width=.7\textwidth]{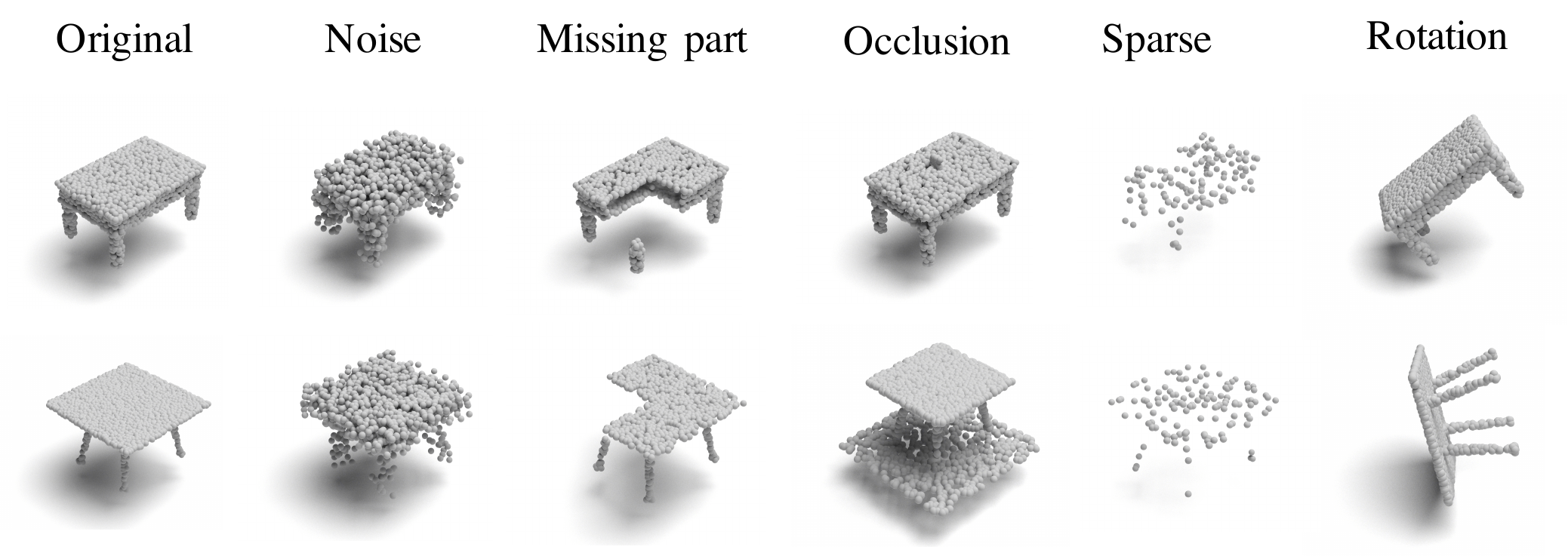}
     \caption{A few samples of the RobustPointSet dataset with different transformations.}
     \label{fig:datsaset}
 \end{figure*}

We present RobustPointSet, a dataset designed to evaluate and benchmark the robustness of pointcloud classifiers to transformations ``unseen'' at train time.
RobustPointSet is based on ModelNet40~\cite{wu20153d}, contains $12,308$ CAD models of $40$ different categories. In addition to original train and test sets of size $9,840$ and $2,468$, respectively found in ModelNet40, RobustPointSet contains $6$ additional ``transformed'' versions of the original training and test sets.
These sets are used to train and evaluate models under two different strategies detailed in subsections~\ref{s1} and \ref{s2}. As shown in Figure~\ref{fig:datsaset}, the transformations we applied in RobustPointSet include:  

\begin{compactitem}
    \item \textbf{Noise}: We add random noise sampled from a normal distribution to each point. 
    \item \textbf{Missing part}: We split each object to 8 axis-aligned chunks and randomly delete one.
    \item \textbf{Occlusion}: Given an object, we add a random wall and/or a random part from another object to it. The random part is taken similar to `Missing part' set, while walls are points sampled from a plane added to the top, bottom, left, right, front, or back of the object.
    \item \textbf{Sparse}: We randomly set $\sim 93\%$ of the 2048 points of each object to zero i.e., each object has only 128 valid points in this set. 
    \item \textbf{Rotation}: We apply a random rotation on an arbitrary axis to each object.  
    \item \textbf{Translation}: Each object is translated by a 3D vector $t$ sampled from the standard distribution which is normalized as  $t = \alpha \times (\frac {t} {\|t\|})$, where $\alpha = 0.1$.

\end{compactitem}

For all the transformations above, we ensure both training and test sets remain in the normalized range of $[-1, 1]$ along each axis by clipping. 

\section{Baseline Experiments}
We designed two strategies to train and evaluate two groups of general and rotation-invariant models using our RobustPointSet dataset. For all the methods, we used the default hyper-parameters as reported in their papers. To assess the extent to which a model is robust to different input transformations, \textit{we do not apply any data augmentation during training} in any of the two strategies explained in subsections~\ref{s1} and~\ref{s2}. We encourage researchers to follow the same setup. 

We consider 10 different point cloud baseline models under two main groups: PointNet~\cite{qi2017pointnet}, PointNet++~\cite{qi2017pointnet++} (both MSG and SSG), DGCNN~\cite{wang2019dynamic}, DensePoint~\cite{liu2019densepoint}, PointCNN~\cite{li2018pointcnn}, PointMask~\cite{taghanaki2020pointmask}, PointConv~\cite{wu2019pointconv}, and Relation-Shape-CNN~\cite{liu2019relation} as general approaches, while SPHnet~\cite{poulenard2019effective} and PRIN~\cite{you2018prin} as rotation invariant methods.


\subsection{Training-domain validation}\label{s1} 
In this setup, we trained each method with the original training set and selected the best performing checkpoint on the original validation set for each method. We then tested the models with selected checkpoints on the six transformed unseen test sets. This experiment shows the vulnerability of the models trained on original data to unseen input transformations. As reported in Table~\ref{tab:s1}, general classification methods mostly fail on the \texttt{Noise} and \texttt{Sparse} test sets.
Simple rigid transformations like translation and rotation are sufficient to degrade test set classification accuracy by up to 10\% and 80\%, respectively.
Methods that leverage local information suffer greatly, with reduction of up to 85\% in accuracy when the notion of neighborhood is changed in test time through the addition of noise or introduction of sparsity.
Among general approaches, PointNet outperforms others with average accuracy of 61.97\%. Although rotation invariant methods such as SPHnet and PRIN show robustness to rotations and translation, they fail on other transformations with high error rates.

\renewcommand{\arraystretch}{.9}
\begin{table*}[h!] 
\centering 
\setlength{\tabcolsep}{2pt}
\caption{Training-domain validation results on our RobustPointSet test sets. The \texttt{Noise} column for example shows the result of training on the \texttt{Original} train set and testing with the \texttt{Noise} test set. RotInv refers to rotation-invariant models.} 
\label{tab:s1}
\begin{tabular}{clcccccccc}
\toprule
                               Type     & Method             & Original & Noise & Translation & MissingPart & Sparse & Rotation & Occlusion & Avg. \\ \midrule
\multirow{8}{*}{\STAB{\rotatebox[origin=c]{90}{General}}}            & PointNet          & 89.06    & \textbf{74.72} & 79.66       & 81.52        & \textbf{60.53}  & 8.83     & 39.47     & \textbf{61.97}   \\
                                    & PointNet++ (M)   & 91.27    & 5.73  & 91.31       & 53.69        & 6.65   & 13.02    & 64.18     & 46.55   \\
                                    & PointNet++ (S)  & 91.47    & 14.90 & 91.07       & 50.24        & 8.85   & 12.70    & 70.23     & 48.49   \\
                                    & DGCNN             & \textbf{92.52}    & 57.56 & \textbf{91.99}       & 85.40        & 9.34   & 13.43    & \textbf{78.72}     & 61.28   \\
                                    & PointMask         & 88.53    & 73.14 & 78.20       & 81.48        & 58.23  & 8.02     & 39.18     & 60.97   \\
                                    & DensePoint         & 90.96    & 53.28 & 90.72       & 84.49        & 15.52  & 12.76    & 67.67     & 59.40   \\
                                    & PointCNN         & 87.66    & 45.55 & 82.85       & 77.60        & 4.01    & 11.50    & 59.50     & 52.67   \\
                                    & PointConv                       & 91.15    & 20.71 & 90.99       & 84.09        & 8.65  & 12.38     & 45.83     & 50.54 \\
                                    & R-ShapeCNN & 91.77    & 48.06 & 91.29       & \textbf{85.98}        & 23.18  & 11.51    & 75.61     & 61.06   \\ \midrule
\multirow{2}{*}{\STAB{\rotatebox[origin=c]{90}{RInv}}} & SPHnet             & 79.18    & 7.22  & \textbf{79.18}       & 4.22         & 1.26   & 79.18    & 34.33     & 40.65   \\
                                    & PRIN               &   73.66       &   30.19    &    41.21          &   44.17           &  4.17       &   68.56        &      31.56     & 41.93        \\ \bottomrule
\end{tabular}
\end{table*}

\subsection{Leave-one-out validation}\label{s2} 
This experiment shows whether data augmentation helps in generalizing to an unseen transformation. Here, we trained the models with six training sets and tested on the \textit{test set} of the held out set. For example, consider three sets $\{A_{\{tr,ts\}}, B_{\{tr,ts\}}, C_{\{tr,ts\}}\}$ where $tr$ and $ts$ refer to training and testing sets, respectively. we first train a model using  $\{A_{tr}, B_{tr}\}$ and test on $C_{ts}$. Next, we train with $\{A_{tr}, C_{tr}\}$ and test on $B_{ts}$. Finally, we train with $\{B_{tr}, C_{tr}\}$ and test on $A_{ts}$. This experiments shows whether different transformations help to generalize to an unseen transformation.

\renewcommand{\arraystretch}{0.9}
\begin{table*}[h!]
\centering
\setlength{\tabcolsep}{2pt}
\caption{Leave-one-out validation strategy classification results on our RobustPointSet test sets. For example, the \texttt{Noise} column shows the result of training on \texttt{\{Original, Translation, Missing part, Sparse, Rotation, Occlusion\}} train sets and testing with the \texttt{Noise} test set. RotInv refers to rotation-invariant models.}
\label{tab:s2}
\begin{tabular}{clcccccccc}
\toprule
    Type &                                    Method &  Original &  Noise &  Translation &  MissingPart &  Sparse &  Rotation &  Occlusion &  Avg. \\
\midrule
\multirow{8}{*}{\STAB{\rotatebox[origin=c]{90}{General}}} &             PointNet &     88.35 &	72.61 &	81.53 &	82.87 &	\textbf{69.28} &	9.42 &	35.96 &	\textbf{62.86} \\
         &   PointNet++ (M) &     91.55 &	50.92 &	91.43 &	77.16 &	16.19 &	12.26 &	\textbf{70.39} &	58.56  \\
         &   PointNet++ (S) &     91.76 &	49.33 &	91.10 &	78.36 &	16.72 &	11.27 &	68.33 &	58.12 \\
         &               DGCNN &     \textbf{92.38} &	66.95 &	91.17 &	85.40 &	6.49 &	14.03 &	68.79 &	60.74 \\
         &    PointMask &     88.03 &	\textbf{73.95} &	80.80 &	82.83 &	63.64 &	8.97 &	36.69 &	62.13  \\
         &      DensePoint  &     91.00 &	42.38 & 90.64 &	85.70 &	20.66 &	8.55 &	47.89 &	55.26 \\
         &             PointCNN &     88.91 &	73.10 &	87.46 &	82.06 &	7.18 &	13.95 &	52.66 &	57.90 \\
         &           PointConv &     91.07 &	66.19 &	\textbf{91.51} &	84.01 &	19.63 &	11.62 &	44.07 &	58.30 \\
         &  R-Shape-CNN &     90.52 &	36.95 &	91.33 &	\textbf{85.82} &	24.59 &	8.23 &	60.09 &	56.79 \\ \midrule
         \multirow{2}{*}{\STAB{\rotatebox[origin=c]{90}{RInv}}} &       SPHnet &     79.30 &	8.24 &	76.02 &	17.94 &	6.33 &	\textbf{78.86} &	35.96 &	43.23 \\
         &                    PRIN &     76.54 &	55.35 &	56.36 &	59.20 &	4.05 &	73.30 &	36.91 &	51.67\\
\bottomrule
\end{tabular}
\end{table*}

We observe from the results in Table~\ref{tab:s2} that minor improvements of upto 10\% on average are obtained on a few methods, while others remain relatively unaffected.
The improvement could be attributed to the use of different transformations during training, which forces the model to choose a robust set of features to make predictions.
However, it is clear that none of the methods are close to their accuracy on the ``Original" test set. 

\section{Conclusion}
We have presented RobustPointSet, a dataset for studying the  robustness of point cloud-based neural networks.
We benchmarked 10 state-of-the-art models and showed that none of the models, particularly the ones that exploit local neighborhood information, do well on transformed test sets.
Our results show that while PointNet~\cite{qi2017pointnet} and DGCNN~\cite{wang2019dynamic} perform well on average, there is no single approach that consistently performs better than the others across a range of data transformations.
This strongly highlights the need to consider robustness as a metric when evaluating models. We hope our dataset will facilitate future research on robustness analysis for point cloud classification models.
While software frameworks for pointcloud processing are increasingly being developed~\cite{kaolin, torchpoints3d, pytorch3d}, the datasets used for training pointcloud classifiers have largely remaind the same.
RobustPointSet addresses this gap, and we experimentally show the need for developing models robust to data transformations -- particularly those unseen at train time.
In future, we are interested in exploring the design of robust pointcloud classifiers that mitigate the effects introduced by bias in the datasets used to train them.




\bibliography{iclr2021_conference}
\bibliographystyle{iclr2021_conference}


\end{document}